\documentclass[letterpaper]{article}
\usepackage[preprint]{aaai2027}

\usepackage[hyphens]{url}
\usepackage{graphicx}
\def\UrlFont{\rm}
\usepackage{natbib}
\usepackage{caption}
\usepackage{booktabs}
\usepackage{amsmath}
\usepackage{amssymb}
\usepackage{array}
\usepackage{listings}
\usepackage{chngcntr}
\usepackage{dblfloatfix}
\usepackage{fontawesome5}

\usepackage[
  colorlinks=true,
  linkcolor=blue!55!black,
  citecolor=blue!55!black,
  urlcolor=blue!55!black,
  pdfborder={0 0 0},
  bookmarksnumbered=true,
  bookmarksopen=true
]{hyperref}
\makeatletter
\expandafter\let\expandafter\DARHyperrefVersion\csname ver@hyperref.sty\endcsname
\expandafter\let\csname ver@hyperref.sty\endcsname\relax
\AtBeginDocument{%
  \expandafter\let\csname ver@hyperref.sty\endcsname\DARHyperrefVersion
}
\makeatother

\definecolor{codeblue}{RGB}{0,0,180}
\definecolor{codegreen}{RGB}{0,115,35}
\definecolor{codered}{RGB}{185,25,25}
\definecolor{codegray}{RGB}{105,105,105}
\lstdefinestyle{darpython}{
  language=Python,
  basicstyle=\ttfamily\scriptsize,
  keywordstyle=\color{codeblue},
  commentstyle=\color{codegreen},
  stringstyle=\color{codered},
  numbers=left,
  numberstyle=\tiny\color{codegray},
  numbersep=7pt,
  frame=single,
  framerule=0.3pt,
  rulecolor=\color{codegray},
  xleftmargin=1.8em,
  framexleftmargin=1.4em,
  columns=fullflexible,
  keepspaces=true,
  showstringspaces=false,
  breaklines=true,
  tabsize=4,
  aboveskip=0pt,
  belowskip=0pt
}

\hypersetup{
  pdftitle={Dual Attention Residuals},
  pdfauthor={Xingda Yu; Yining Li; Xinzhang Liu; Zhihao Yang; Haowei He; Chao Wang; Yongxiang Li; Shuangyong Song}
}

\title{Dual Attention Residuals}
\author{
Xingda Yu\textsuperscript{\rm 1,2}\equalcontrib,
Yining Li\textsuperscript{\rm 3,1}\equalcontrib,
Xinzhang Liu\textsuperscript{\rm 1},
Zhihao Yang\textsuperscript{\rm 1},\\
Haowei He\textsuperscript{\rm 1},
Chao Wang\textsuperscript{\rm 1},
Yongxiang Li\textsuperscript{\rm 1}\corresponding,
Shuangyong Song\textsuperscript{\rm 1}\corresponding
}
\affiliations{
\textsuperscript{\rm 1}TeleAI\\
\textsuperscript{\rm 2}Institute of Computing Technology, Chinese Academy of Sciences\\
\textsuperscript{\rm 3}Qiuzhen College, Tsinghua University\\
\href{https://github.com/babyinsunshine/Dual-Attention-Residuals}{%
  \textcolor{black}{\raisebox{-0.08em}{\faGithub}}\,%
  \textcolor{blue}{\texttt{Teleai/Dual-Attention-Residuals}}%
}
}

\begin{document}
\maketitle

\def\DARCombined{1}
\ifdefined\DARCombined
\let\DARMaybeEnd\relax
\else
\documentclass[letterpaper]{article} 
\usepackage[submission]{aaai2027}  
\usepackage[hyphens]{url}  
\usepackage{graphicx} 
\urlstyle{rm} 
\def\UrlFont{\rm}  
\usepackage{natbib}  
\usepackage{caption} 
\frenchspacing  
\usepackage{booktabs}
\usepackage{amsmath}
\usepackage{amssymb}
\usepackage{array}
\pdfinfo{
/TemplateVersion (2027.1)
}
\setcounter{secnumdepth}{0}

\title{Dual Attention Residuals}
\author{Anonymous Submission}
\affiliations{}

\begin{document}
\maketitle
\def\DARMaybeEnd{\end{document}}
\fi
\begin{abstract}
Recent work extends Transformer residual pathways along two complementary axes: historical retrieval provides direct access to earlier computation, whereas multi-stream methods maintain multiple evolving representations.
These directions have largely developed independently, leaving unexplored how depth-wise retrieval and multi-stream representation can complement each other within a unified residual architecture.
We propose \emph{Dual Attention Residuals} (DAR), which combines these capabilities by allowing the two trajectories to participate reciprocally in depth retrieval.
For each target stream, DAR computes depth weights from normalized states in the opposite stream and applies them to values from the target stream's own history.
The retrieved states are combined as input to a Transformer branch. To control overhead, DAR retrieves over completed block histories while maintaining two temporary stream states within each block.
Experiments on dense models from 0.1B to 1B parameters and a 7B sparse-MoE model demonstrate the effectiveness of DAR across model scales.
Parameter-matched ablations show that DAR benefits from both its dual-stream representation and reciprocal cross-stream retrieval.
Representation and intervention analyses further show that reciprocal cross-stream selection preserves depth-wise diversity and avoids the redundancy or functional imbalance observed in alternative two-stream designs.
\end{abstract}

\section{Introduction}

Residual connections form the main pathway for transporting representations across Transformer depth~\citep{he2016resnet,vaswani2017attention}.
In a standard Transformer, successive computational branches are accumulated into a single residual stream and propagated sequentially through the network.
Recent work extends this pathway along two complementary directions.
Historical retrieval methods expand access across depth: Transparent Attention and DenseFormer expose earlier representations to later computation~\citep{bapna2018transparent,pagliardini2024denseformer}, while Attention Residuals makes this access selective through learned depth-wise retrieval~\citep{chen2026attentionresiduals}.
Multi-stream residual methods instead expand the pathway into multiple learned trajectories, beginning with Hyper-Connections~\citep{zhu2024hyperconnections} and extending to Frac-Connections, MUDDFormer, and mHC~\citep{zhu2025fracconnections,xiao2025muddformer,xie2025mhc}.
Historical retrieval introduces freedom over which depths to access, whereas multi-stream residuals introduce freedom over which residual trajectories to maintain and combine.
These two degrees of freedom have been studied largely in isolation, leaving open whether they can interact productively within historical retrieval.

A straightforward combination is to equip each residual stream with an independent historical retriever.
This, however, produces only parallel instances of single-stream retrieval: each trajectory determines which depths to access and supplies the content returned from those depths, without allowing the other trajectory to influence its selection.
We call this use of the same trajectory both to score depths and to supply the retrieved content \emph{address-content coupling}.
To bring multi-stream flexibility into depth-wise retrieval, the trajectories must instead interact during selection.
This makes trajectory diversity part of the retrieval rule: historical content in one stream can receive a high depth weight because the other stream supplies a matching address, even when the content representation would not itself strongly match the target query.
The trajectories therefore contribute not only different stored content but also complementary signals for selecting it.

To make multi-stream interaction part of depth-wise retrieval, we introduce Dual Attention Residuals (DAR) with reciprocal cross-stream addressing.
For each target stream, DAR computes depth weights from normalized states in the opposite stream and applies them to values from the target stream's history.
The resulting stream-specific states are mixed as input to an unchanged Transformer branch.
DAR then updates both residual trajectories through gated branch writes, with constrained mixing transporting the existing partial state within each block.
For efficient scaling, Block DAR retrieves from completed block histories at a coarser depth granularity.

Our contributions are summarized as follows:
\begin{itemize}
  \item We show that independently evolving residual trajectories can serve not only as parallel information pathways, but also as complementary signals for selecting historical depths.
  \item We propose DAR, which realizes this interaction through reciprocal cross-stream keys and self-stream values while leaving the Transformer branches unchanged.
  \item We validate DAR through scaling experiments on dense models from 0.1B to 1B parameters and a 7B sparse-MoE model, retrieval ablations, and representation analyses, demonstrating consistent improvements over standard residual connections and Attention Residuals.
\end{itemize}

\section{Related Work}

\subsection{Residual Transport and Multi-Stream Topologies}

Standard residual connections transport a single representation through network depth~\citep{he2016resnet,vaswani2017attention}.
Transformer normalization improves optimization while preserving this single residual trajectory~\citep{ba2016layernorm,xiong2020layernorm,zhang2019rmsnorm}.
Hyper-Connections instead expand the residual pathway into multiple streams and learn how Transformer branches read from and write to them~\citep{zhu2024hyperconnections}.
Frac-Connections and MUDDFormer further enrich cross-layer connectivity, while mHC constrains multi-stream mixing to preserve stable transport at scale~\citep{zhu2025fracconnections,xiao2025muddformer,xie2025mhc}.
These methods show how multiple residual streams can transport and mix information within a layer.
DAR extends this idea across depth by retaining both trajectories in the historical state bank and applying depth-wise retrieval over them.

\subsection{Depth-Wise Historical Retrieval}

Another line of work makes representations from earlier depths directly accessible to later computation.
Transparent attention and DenseFormer aggregate intermediate layer states, while Attention Residuals (AttnRes) uses learned pseudo-queries to perform softmax retrieval over previous residual states~\citep{bapna2018transparent,pagliardini2024denseformer,chen2026attentionresiduals}.
DeepCrossAttention also learns input-dependent combinations of earlier layer outputs through depth-wise cross-attention~\citep{heddes2025deepcrossattention}.
Delta Attention Residuals replaces cumulative hidden-state candidates with individual branch outputs~\citep{luo2026deltaattention}.
These methods vary the candidate states and aggregation rules. In single-stream historical retrieval, however, the representations that determine depth weights and the values aggregated under them still come from the same residual trajectory.
DAR instead uses the opposite trajectory to determine each value's depth weight while keeping that value in its original stream.

\section{Preliminaries}

We consider a Pre-Norm decoder-only Transformer and treat self-attention and feed-forward modules as separate residual branches.
We index these branches from $\ell=1$, with $h_1$ denoting the token embedding.
Let $h_\ell\in\mathbb{R}^{T\times d}$ be the input to branch $\ell$, and let $f_\ell$ denote the complete branch transformation, including its input normalization.
A standard residual branch updates $h_{\ell+1}=h_\ell+f_\ell(h_\ell)$, where $T$ and $d$ are the sequence length and hidden size.
Depth retrieval is token-wise, so we suppress the token index.

\paragraph{Attention Residuals.}
AttnRes~\citep{chen2026attentionresiduals} replaces cumulative addition with softmax retrieval over the token embedding and preceding branch outputs.
Following its original formulation, we use the positive kernel
\begin{equation}
  \phi(q,k)=\exp\!\left(q^\top\mathrm{Norm}_h(k)\right)
\end{equation}
and write the depth-attention weights as
\begin{equation}
  \pi_{r\to\ell}
  =
  \frac{\phi(q_\ell,k_r)}
       {\sum_{u=0}^{\ell-1}\phi(q_\ell,k_u)}.
\end{equation}
For every residual branch $\ell$, the query, keys, and values are
\begin{equation}
  q_\ell=w_\ell,
  \qquad
  k_r=v_r=
  \begin{cases}
    h_1, & r=0, \\
    f_r(h_r), & 1\le r\le \ell-1,
  \end{cases}
\end{equation}
where $w_\ell\in\mathbb{R}^d$ is a branch-specific learnable vector and $\mathrm{Norm}_h$ denotes RMSNorm along the hidden dimension.
Thus $v_0=h_1$ is the embedding candidate, while $v_r=f_r(h_r)$ for $r\ge1$ is the output of an earlier residual branch.
The input to branch $\ell$ is
\begin{equation}
  h_\ell
  =\sum_{r=0}^{\ell-1}\pi_{r\to\ell}v_r.
\end{equation}
The same source $v_r$ therefore determines its depth weight and supplies its value, which we call \emph{address-content coupling}.

\section{Methods}

\begin{figure*}[t]
\centering
\includegraphics[width=\linewidth]{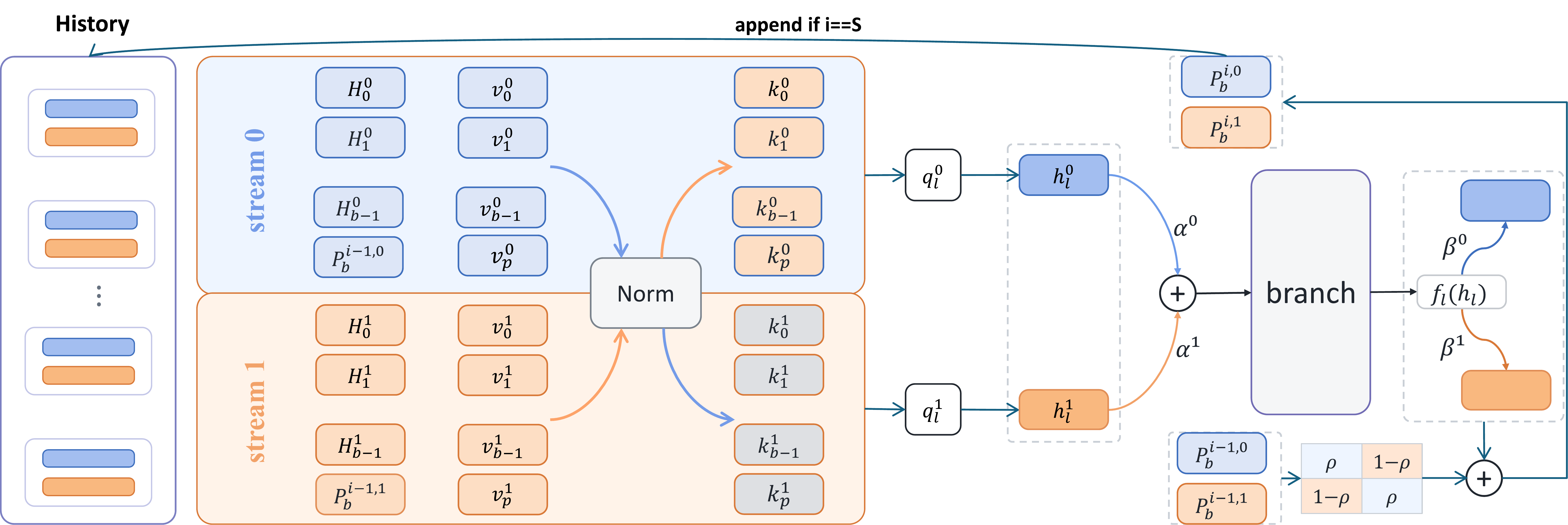}
\caption{Block DAR, shown for branch $i>1$ in block $b$. Opposite-stream keys retrieve self-stream values, which are mixed for an unchanged Transformer branch; its gated output combines with constrained transport of $P_b^{i-1}$ to form $P_b^i$. The partial path is omitted for $i=1$, and $P_b^S$ is appended as $H_b$ at the block boundary.}
\label{fig:dar}
\end{figure*}

DAR combines depth-wise historical retrieval with multi-stream residual pathways, allowing the two trajectories to interact reciprocally during depth selection rather than operating as independent retrievers.
As shown in Figure~\ref{fig:dar}, each stream uses the other to weight its own history.
The results are combined as input to an attention or feed-forward branch, whose output is written back to both streams.

\subsection{Overview and Notation}

DAR maintains an ordered history bank $\mathcal H$.
Its $r$-th history state is
\begin{equation}
  H_r=[v_r^0,v_r^1],
  \qquad
  v_r^j\in\mathbb{R}^{T\times d},\quad j\in\{0,1\},
\end{equation}
We initialize $H_0=[v_0^0,v_0^1]=[h_1,h_1]$ from the token embedding.

Block DAR partitions the residual branches into consecutive blocks.
Let $b\ge1$ index blocks and let the \emph{block size} $K$ denote the number of Transformer layers per block.
Because each layer contains an attention branch and a feed-forward branch, a block contains $S=2K$ residual branches; we use $i\in\{1,\ldots,S\}$ to index their positions within the block.
Setting $K=1$ ($S=2$) makes each Transformer layer an independent block, which we call Full DAR.
We explain this layer-level definition in \emph{Full DAR at Layer Granularity}.

\subsection{Block Dual-Stream Retrieval}

Before block $b$, the history bank contains
\begin{equation}
  \mathcal H_b=(H_0,H_1,\ldots,H_{b-1}).
\end{equation}
Within block $b$, DAR maintains two temporary stream states across its residual branches.
We denote their values after the first $i$ branches by $P_b^i=[P_b^{i,0},P_b^{i,1}]$.
$P_b^1$ is constructed from the first branch output, whereas $P_b^i$ for $i>1$ follows the update rule defined in the next subsection.
These states remain local to block $b$ and enter the history bank only at the block boundary.
For branch $\ell$ at position $i$ in block $b$, we define its retrieval set as
\begin{equation}
  \mathcal C_\ell
  =
  \begin{cases}
    \mathcal H_b, & i=1, \\
    (\mathcal H_b,P_b^{i-1}), & i>1.
  \end{cases}
\end{equation}
Only branches with $i>1$ include the current partial state in retrieval.

Each candidate $C_r=[C_r^0,C_r^1]\in\mathcal C_\ell$ contains one state from each stream.
For output stream $j$, DAR uses the opposite-stream state $C_r^{1-j}$ as the key and the same-stream state $C_r^j$ as the value:
\begin{equation}
  k_r^j=C_r^{1-j},
  \qquad
  v_r^j=C_r^j,
  \qquad j\in\{0,1\}.
\end{equation}
Every residual branch and stream has an independent query $q_\ell^j\in\mathbb{R}^d$.
The resulting depth weights and retrieved stream states are
\begin{align}
  \pi_{r\to\ell}^j
  &=
  \frac{\phi(q_\ell^j,k_r^j)}
       {\sum_{C_u\in\mathcal C_\ell}
        \phi(q_\ell^j,k_u^j)}, \\
  h_\ell^j
  &=
  \sum_{C_r\in\mathcal C_\ell}
  \pi_{r\to\ell}^jv_r^j.
\end{align}
The opposite stream therefore controls where stream $j$ retrieves from, while the aggregated content remains in stream $j$'s own residual space.

\subsection{Block-Local Mixing and Gated Writes}

Following mHC~\citep{xie2025mhc}, DAR predicts token-wise mixing and write coefficients from the two retrieved states:
\begin{align}
  \bar h_\ell
  &=\mathrm{Norm}([h_\ell^0;h_\ell^1]), \\
  \alpha_\ell
  &=\sigma(\bar h_\ell W_{\alpha,\ell}+b_{\alpha,\ell}), &
  \beta_\ell
  &=2\sigma(\bar h_\ell W_{\beta,\ell}+b_{\beta,\ell}),
\end{align}
where $\alpha_\ell\in(0,1)^2$ and $\beta_\ell\in(0,2)^2$.
The unchanged Transformer branch receives
\begin{equation}
  h_\ell=\alpha_\ell^0h_\ell^0+\alpha_\ell^1h_\ell^1
\end{equation}
and produces $u_\ell=f_\ell(h_\ell)$.

Later branches in a block transport the existing partial state with
\begin{equation}
\begin{aligned}
  \bar p_{b,i}
  &=
  \mathrm{Norm}([P_b^{i-1,0};P_b^{i-1,1}]), \\
  \rho_{b,i}
  &=
  \sigma(\bar p_{b,i}W_{\rho,\ell}+b_{\rho,\ell}), \\
  M(\rho_{b,i})
  &=
  \begin{bmatrix}
    \rho_{b,i} & 1-\rho_{b,i} \\
    1-\rho_{b,i} & \rho_{b,i}
  \end{bmatrix}.
\end{aligned}
\end{equation}
For two streams, this parameterization directly guarantees a doubly stochastic matrix: each row and column sums to one, so no iterative normalization is required.
The complete partial-state update is
\begin{equation}
  P_b^i
  =
  \begin{cases}
    [\beta_\ell^0u_\ell,\beta_\ell^1u_\ell], & i=1, \\[2pt]
    M(\rho_{b,i})P_b^{i-1}
    +[\beta_\ell^0u_\ell,\beta_\ell^1u_\ell], & i>1.
  \end{cases}
\end{equation}
Thus the first branch initializes the partial state, while subsequent branches mix its two streams before adding their gated writes.
At the block boundary, $P_b^S$ is stored as the new history state
\begin{equation}
  H_b=P_b^S.
\end{equation}
At the network output, learned output queries apply the same cross-key/self-value retrieval to the history bank.
The sum $h_{\mathrm{out}}^0+h_{\mathrm{out}}^1$ is passed to the original final normalization and vocabulary projection.
\ifdefined\DARCombined
Appendix Figure~\ref{fig:block-dar-pseudocode} provides framework-level pseudocode for the complete Block DAR forward pass.
\else
Figure~1 in the supplementary material provides framework-level pseudocode for the complete Block DAR forward pass.
\fi

\subsection{Full DAR at Layer Granularity}

The finest meaningful DAR block is one Transformer layer, $K=1$, which contains an attention branch and a feed-forward branch ($S=2$).
We call this layer-wise construction Full DAR because every Transformer layer contributes a separate history state.

A branch-wise construction with $S=1$ is not used.
With only the branch output $u_\ell$, its history state would reduce to
\begin{equation}
  H_\ell
  =[\beta_\ell^0u_\ell,\beta_\ell^1u_\ell].
\end{equation}
Consequently, at every token position $p$,
\begin{equation}
  H_{\ell,p}^1
  =\frac{\beta_{\ell,p}^1}{\beta_{\ell,p}^0}H_{\ell,p}^0.
\end{equation}
The two stored streams are therefore collinear and differ only by a positive scalar.
They cannot encode distinct summaries of the branch computation: the cross-stream key and self-stream value both originate from the same underlying content direction.
Independent stream queries may still produce different readouts over depth, but they do not remove the degeneracy of each history state.
Moreover, with no preceding partial state, the constrained transport $M(\rho)$ is never applied.

\subsection{Design Comparisons}

\paragraph{AttnRes + VProj.}
This single-stream control assigns each stored state an independent value projection.
\begin{equation}
  \tilde v_r=W_{v,r}v_r .
\end{equation}
VProj therefore adds one $d\times d$ map to each branch's value path, totaling $O(Rd^2)$ parameters for $R$ sources.
A source-specific key projection is algebraically equivalent to replacing $q_\ell$ with $q_{\ell,r}=W_{k,r}^\top q_\ell$, because the key map can be moved to the query side of the dot product.
It also introduces the same number of additional parameters as VProj.

\paragraph{Retrieval variants.}
We compare four parameter-matched retrieval rules that differ only in the key and value sources for output stream $j$:
\begin{equation}
\begin{aligned}
\text{DAR-SelfKV:}\quad
  &(k_r^j,v_r^j)=(C_r^j,C_r^j),\\
\text{DAR-FixedKV:}\quad
  &(k_r^j,v_r^j)=(C_r^0,C_r^1),\\
\text{DAR-CrossV:}\quad
  &(k_r^j,v_r^j)=(C_r^j,C_r^{1-j}),\\
\text{DAR:}\quad
  &(k_r^j,v_r^j)=(C_r^{1-j},C_r^j).
\end{aligned}
\end{equation}
\textbf{DAR-SelfKV.}
Each stream retrieves independently from its own history, using each state both to score its depth and to supply the retrieved content; DAR-SelfKV thus retains AttnRes's within-trajectory address-content coupling.
Beyond this coupling, the streams interact only after retrieval through branch-input mixing and subsequent writes, so neither trajectory can help the other decide which historical depths are relevant.
This limited interaction can underuse complementary information carried by the other trajectory and encourage similar retrieval behavior across streams.

\textbf{DAR-FixedKV.}
Both outputs use stream~0 as the key source and stream~1 as the value source, with their independent queries producing different depth weights over the same two banks.
This fixed assignment makes the interaction one-sided: stream~0 supplies the key features used to determine the depth weights for both outputs, whereas stream~1 supplies all retrieved values.
The two trajectories never exchange these roles.

\textbf{DAR-CrossV.}
Each stream scores its own history but applies the resulting weights to values from the opposite stream.
Although this exchanges historical content, the retrieved state indexed by stream $j$ no longer comes from stream $j$'s own value trajectory.
Repeatedly crossing the value path can therefore weaken the continuity of stream-specific content, rather than using one trajectory to evaluate the history retained by the other.

\textbf{DAR.}
For output stream $j$, the opposite trajectory supplies the keys while stream $j$ retains its own values.
The retrieved content therefore remains in its original residual trajectory, while the other trajectory contributes a complementary signal for selecting its relevant depths.
Applying the rule symmetrically in both directions lets each stream serve as both a content trajectory and an address for the other, enabling reciprocal interaction during historical retrieval.

\section{Experiments}

We evaluate DAR under matched training budgets across dense models from 0.1B to 1B parameters and a 7B sparse-MoE model, comparing against standard residual connections, Attention Residuals, and mHC-4. Beyond validation loss, we report training efficiency under matched fused implementations, conduct parameter-matched ablations of retrieval rules and value transformations, and analyze the learned streams through historical retrieval patterns, cross-depth representation similarity, and intervention-based output recovery.

\subsection{Experimental Setup}

\paragraph{Models and data.}
We evaluate three smaller dense models with 0.1B, 0.3B, and 0.5B non-vocabulary parameters, a main 1B dense model, and a sparse-MoE model.
The smaller models are trained on FineWeb-Edu-10B~\citep{penedo2024fineweb} with a token budget equal to $20\times$ their parameter count and evaluated on 10M held-out tokens.
The 1B model is trained on 50B FineWeb-Edu-100B tokens.
The sparse model has 7.22B total parameters and activates 1.40B per token, including vocabulary parameters; it is trained for 60K steps, corresponding to 31.46B FineWeb-Edu-100B tokens.
All runs use a context length of 4096, with global batch sizes of 256 for dense models and 128 for sparse MoE.

\paragraph{Training and implementation.}
All methods within each setting share the same data, backbone, and training configuration.
The mHC-4 baseline uses four streams; to facilitate analysis, we additionally train mHC-2 under the same 0.5B configuration.
Models are trained on a single node with eight H800 GPUs using AdamW~\citep{loshchilov2019adamw}.
Dense experiments use PyTorch with DeepSpeed~\citep{rasley2020deepspeed} and torch.compile~\citep{ansel2024pytorch2}, whereas sparse-MoE experiments use Megatron-Core~\citep{shoeybi2019megatron} and fused AttnRes/DAR operator implementations that are not yet fully optimized.
For the parameter-matched 0.5B ablation, we set the FFN width of AttnRes + VProj to 4992.
We use Block DAR in the main dense and sparse-MoE experiments and evaluate both Block and Full DAR in the smaller models.
\ifdefined\DARCombined
Detailed configurations and parameter counts are provided in Appendix Tables~\ref{tab:main-hparams}--\ref{tab:moe-training}.
\else
Detailed configurations and parameter counts are provided in Tables~1--6 of the supplementary material.
\fi

\subsection{Main Results}

\paragraph{1B-parameter setting.}
We first assess the 1B models using FineWeb-Edu validation loss and Paloma token PPL, byte PPL, and BPB~\citep{magnusson2024paloma}.
Table~\ref{tab:main-quality} compares the baseline, block-form Attention Residuals, and block-form DAR.

\begin{figure}[t]
\centering
\includegraphics[width=\linewidth]{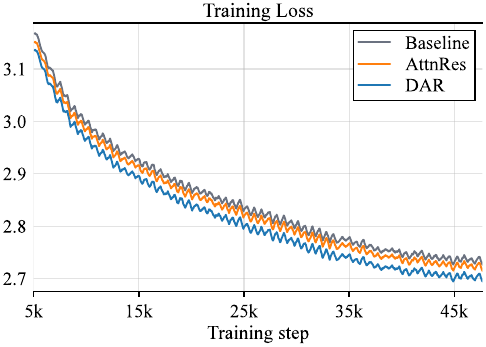}
\caption{Training loss of the 1B dense models (300-step moving average).}
\label{fig:main-1b-loss}
\end{figure}

\begin{table}[t]
\centering
\small
\begin{tabular}{@{\hspace{3.2pt}}l@{\hspace{6.4pt}}c@{\hspace{6.4pt}}c@{\hspace{6.4pt}}c@{\hspace{6.4pt}}c@{\hspace{3.2pt}}}
\toprule
Method & Val. loss & Token PPL & Byte PPL & BPB \\
\midrule
Baseline & 2.716 & 21.575 & 2.040 & 1.029 \\
AttnRes & 2.704 & 20.707 & 2.021 & 1.015 \\
DAR & \textbf{2.684} & \textbf{20.264} & \textbf{2.011} & \textbf{1.008} \\
\bottomrule
\end{tabular}
\caption{Language-model fit for the 1B models. Paloma reports token perplexity (PPL), byte PPL, and bits per byte (BPB); aggregate PPL/BPB values are micro-averaged over all tokens or bytes from the 16 sources. Lower is better for all columns.}
\label{tab:main-quality}
\end{table}

DAR performs best on all four reported measures and obtains the lowest perplexity on 15 of the 16 Paloma domains; Attention Residuals is best only on 4chan.
\ifdefined\DARCombined
The complete domain results are provided in Appendix~\ref{app:1b-evaluation}.
\else
The complete domain results are provided in the section ``Detailed 1B Evaluation'' of the supplementary material.
\fi
Figure~\ref{fig:main-1b-loss} further shows that DAR maintains lower smoothed training loss than both baselines for most of training.

To test whether these gains extend beyond language modeling to downstream task performance, we evaluate the same checkpoints zero-shot with the default lm-evaluation-harness task formulations~\citep{biderman2024reproducible}.
The eight tasks are HellaSwag~\citep{zellers2019hellaswag}, PIQA~\citep{bisk2020piqa}, ARC-Easy~\citep{clark2018arc}, WinoGrande~\citep{sakaguchi2020winogrande}, BoolQ~\citep{clark2019boolq}, OpenBookQA~\citep{mihaylov2018openbookqa}, SocialIQA~\citep{sap2019socialiqa}, and SciQ~\citep{welbl2017sciq}.
Table~\ref{tab:zero-shot-1b} shows that DAR improves over the baseline on all eight tasks and achieves the best score on five; Attention Residuals is best on PIQA, ARC-Easy, and SciQ.

\begin{table*}[t]
\centering
\small
\begin{tabular}{@{\hspace{2.5pt}}l*{9}{@{\hspace{5pt}}c}@{\hspace{2.5pt}}}
\toprule
Method & HellaSwag & PIQA & ARC-E & WinoGrande & BoolQ & OpenBookQA & SocialIQA & SciQ & Avg. \\
\midrule
Baseline & 53.96 & 72.14 & 49.92 & 53.67 & 61.16 & 33.00 & 40.02 & 72.40 & 54.53 \\
AttnRes & 54.65 & \textbf{73.94} & \textbf{52.10} & 53.83 & 58.84 & 31.40 & 40.84 & \textbf{76.00} & 55.20 \\
DAR & \textbf{55.35} & 73.01 & 51.30 & \textbf{54.22} & \textbf{62.45} & \textbf{33.20} & \textbf{40.94} & 74.60 & \textbf{55.63} \\
\bottomrule
\end{tabular}
\caption{Zero-shot performance (\%) of the 1B models. Higher is better; Avg. is the unweighted mean across tasks.}
\label{tab:zero-shot-1b}
\end{table*}

\paragraph{Other model sizes.}
Table~\ref{tab:size-sweep} compares the residual baseline, mHC-4, and the block and full forms of DAR and Attention Residuals.

\begin{table}[t]
\centering
\begin{tabular}{lccc}
\toprule
Method & 0.1B & 0.3B & 0.5B \\
\midrule
Baseline & 3.577 & 2.866 & 2.796 \\
DAR-Block & \textbf{3.504} & \textbf{2.803} & 2.674 \\
DAR-Full & 3.515 & 2.804 & \textbf{2.653} \\
AttnRes-Block & 3.572 & 2.823 & 2.754 \\
AttnRes-Full & 3.541 & 2.827 & 2.708 \\
mHC-4 & 3.531 & 2.844 & 2.764 \\
\bottomrule
\end{tabular}
\caption{Validation loss in the small-scale dense suite. Model size excludes vocabulary parameters.}
\label{tab:size-sweep}
\end{table}

DAR consistently outperforms both the baseline and mHC-4.
Compared with the baseline, its best variant reduces validation loss by 2.03\%, 2.22\%, and 5.12\% at 0.1B, 0.3B, and 0.5B, respectively.
Block DAR performs best at 0.1B and 0.3B, while Full DAR performs best at 0.5B.

\begin{figure}[t]
\centering
\includegraphics[width=\linewidth]{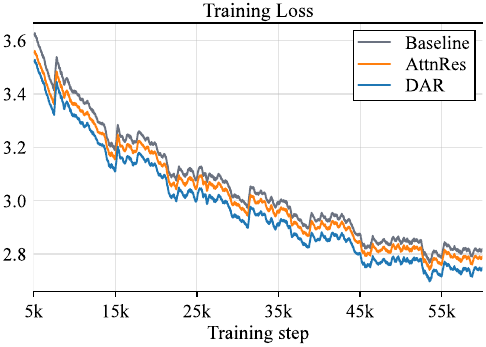}
\caption{Training language-modeling loss of the 7B sparse-MoE models (300-step moving average).}
\label{fig:moe-7b-loss}
\end{figure}

\paragraph{Sparse-MoE setting.}
To test whether DAR remains effective as the model scales up, we further evaluate it on a 7B sparse-MoE model.
DAR achieves the lowest validation loss (2.739), compared with 2.785 for Attention Residuals and 2.814 for the baseline.
Its training curve remained below both baselines for most of training (Figure~\ref{fig:moe-7b-loss}).
The reported values and curves include only the language-modeling loss and exclude the auxiliary router load-balancing loss.
\ifdefined\DARCombined
Appendix Figure~\ref{fig:moe-load-balancing} shows similar router balance across the three models.
\else
Figure~2 in the supplementary material shows similar router balance across the three models.
\fi

\ifdefined\DARCombined
Appendix Figure~\ref{fig:moe-7b-retrieval-maps} compares the historical retrieval distributions learned by the two sparse-MoE DAR streams before the attention and MLP branches.
\else
Figure~3 in the supplementary material compares the historical retrieval distributions learned by the two sparse-MoE DAR streams before the attention and MLP branches.
\fi
This complementary retrieval indicates a division of labor across depth: the streams expose different historical contexts to the same Transformer branch rather than reproducing a single retrieval pattern.

\subsection{Ablation Study}
\label{sec:diagnostics}

\begin{table}[t]
\centering
\begin{tabular}{@{\hspace{3pt}}l@{\hspace{6pt}}c@{\hspace{6pt}}c@{\hspace{6pt}}c@{\hspace{3pt}}}
\toprule
Method & Key source & Value source & Val. loss  \\
\midrule
AttnRes & $v_r$ & $v_r$ & 2.708   \\
AttnRes + VProj & $v_r$ & $\tilde v_r$ & 2.757 \\
DAR-SelfKV & $C_r^j$ & $C_r^j$ & 2.678  \\
DAR-FixedKV & $C_r^0$ & $C_r^1$ & 2.690  \\
DAR-CrossV & $C_r^j$ & $C_r^{1-j}$ & 2.664 \\
DAR & $C_r^{1-j}$ & $C_r^j$ & 2.653  \\
\bottomrule
\end{tabular}
\caption{Key--value retrieval ablations in the 0.5B Full setting under parameter-matched training. The key and value columns correspond to the right-hand side of each retrieval definition in the Design Comparisons subsection.}
\label{tab:kv-ablation}
\end{table}

Table~\ref{tab:kv-ablation} compares the retrieval designs under parameter-matched training.
VProj performs worse than AttnRes, indicating that address--content coupling cannot be effectively addressed by value projection alone.
All multi-stream variants outperform AttnRes, showing that multi-stream residual pathways and historical retrieval are complementary.
Among these variants, DAR achieves the lowest loss, favoring cross-stream keys with self-stream values among the tested retrieval rules.

\subsection{Stream Analysis}

We then use cross-stream CKA and single-stream interventions to examine whether the two streams learn distinct representations and both contribute to predictions.

\begin{figure*}[t]
\centering
\includegraphics[width=0.96\textwidth]{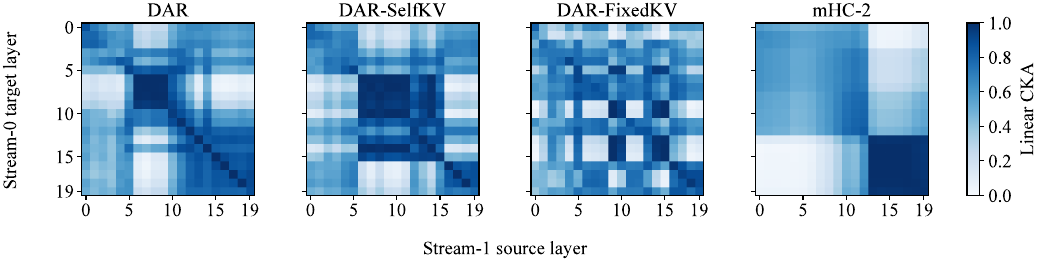}
\caption{Cross-stream linear CKA for post-MLP residual-stream states in the 0.5B models. Rows are stream-0 layers and columns are stream-1 layers; mHC-2 is the two-stream configuration used for this analysis.}
\label{fig:cka-cross-stream}
\end{figure*}

\begin{figure*}[t]
\centering
\includegraphics[width=0.96\textwidth]{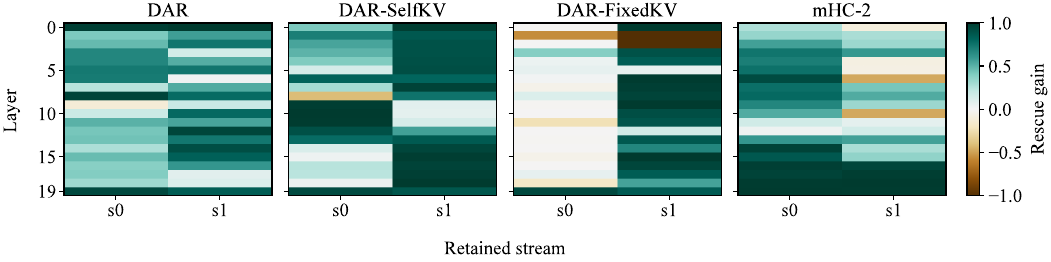}
\caption{Post-MLP residual-stream rescue gain by layer in the 0.5B models. Columns indicate the retained stream; values are normalized against zeroing both streams.}
\label{fig:kl-rescue}
\end{figure*}

\paragraph{Representation similarity between streams.}
We use cross-stream CKA (Figure~\ref{fig:cka-cross-stream}) to test whether the two streams remain distinct rather than duplicate one another across depth.
DAR shows low similarity between the two streams at different depths, while alignment is confined near corresponding late layers, indicating distinct depth-wise trajectories without widespread redundancy.
Using four FineWeb-Edu-10B validation sequences of length 4096, we collect post-MLP stream states over the final 128 tokens and compute cross-stream, cross-depth linear CKA~\citep{kornblith2019cka}.
CKA compares pairwise similarities among token representations; a high value means that token pairs represented similarly at one depth are also represented similarly at the other.

By contrast, DAR-SelfKV's early alignment expands into a broad block across the middle layers, indicating that many different depths in the two streams learn similar representations and thus provide less depth-specific diversity for retrieval.
DAR-FixedKV produces repeated vertical bands, meaning that a few stream-1 depths are similar to many stream-0 depths.
This many-to-one pattern suggests that fixed key--value roles make multiple depths reuse similar representations, reducing the depth-wise diversity available for historical retrieval.
For mHC-2, many pairs of late layers at different depths have high CKA, forming a broad cross-depth region.
This pattern is consistent with a lower effective depth, as representations from distinct late layers become less distinguishable.
\paragraph{Contribution of each stream to predictions.}
To test whether each stream contributes to the model's prediction, we perform layer-wise interventions following~\citet{peng2026ablate}.
Figure~\ref{fig:kl-rescue} shows the desired result for DAR: retaining either stream improves over removing both, but neither stream alone consistently recovers the clean prediction, indicating complementary rather than redundant contributions.
Immediately after the MLP at layer $n$, we separately keep only stream~0, keep only stream~1, or remove both, and then continue the forward pass.
For stream $j$, the rescue gain measures how much keeping it reduces the KL divergence from the clean output relative to removing both:
\[
R_{n,j}=1-\frac{\mathrm{KL}(p_{\mathrm{clean}}\|p_{\mathrm{keep}\text{-}j})}
{\max(\mathrm{KL}(p_{\mathrm{clean}}\|p_{\mathrm{zero}\text{-}both}),10^{-12})}.
\]
Thus, a score near $1$ means that keeping stream $j$ nearly reproduces the clean output, whereas a score near $0$ means that it helps little beyond removing both streams.

In contrast, despite high cross-stream CKA, DAR-SelfKV's KL rescue consistently favors stream~1.
This strong asymmetry indicates that independent self-retrieval does not let the two trajectories cooperate effectively: without DAR's reciprocal cross-stream selection, one stream becomes functionally dominant while the other remains underused.
To explain how this imbalance arises, we examine $\beta_n^j$, which directly controls how much branch output enters stream~$j$.
\ifdefined\DARCombined
Appendix Figure~\ref{fig:selfkv-gate-rescue} shows that SelfKV has a much larger write imbalance than DAR, consistent with its asymmetric rescue.
\else
Figure~4 of the supplementary material shows that SelfKV has a much larger write imbalance than DAR, consistent with its asymmetric rescue.
\fi
DAR-FixedKV's one-sided recovery follows directly from its fixed roles: stream~1 stores all retrievable values, whereas stream~0 supplies only their keys and therefore cannot recover content once stream~1 is removed.
Because stream~0 is never used as retrievable content, this fixed assignment wastes part of the two-stream capacity, consistent with FixedKV's higher validation loss than DAR in Table~\ref{tab:kv-ablation}.
For mHC-2, both streams individually show high late-layer rescue, matching the broad high-CKA region across late depths: because the streams encode similar information, either can replace much of what is lost when the other is removed.
Together, the two analyses indicate redundancy rather than complementary specialization, consistent with a lower effective depth.
Negative stream-1 rescue at several middle layers further shows that this redundancy does not ensure useful cooperation throughout the network.

\subsection{Training Efficiency}

For the 7B sparse-MoE efficiency comparison, Attention Residuals and DAR use a matched Triton prefix-kernel family~\citep{tillet2019triton} following Attention Residuals~\citep{chen2026attentionresiduals}.
The kernels pack normalized key views and values, then compute the retrieved state and its log-partition.
Following the two-phase computation of Attention Residuals, the current partial state is incorporated through an exact online-softmax merge.
DAR evaluates both streams jointly and uses fused residual-read stages to predict $\alpha_\ell$ and $\beta_\ell$, together with $\rho_{b,i}$ for $i>1$, and to form the branch input and optional transported partial state.
The Transformer branch remains unchanged; a post-attention epilogue fuses branch-output handling with the gated stream write.
\ifdefined\DARCombined
More details are provided in Appendix~\ref{app:fused-operators}.
\else
More details are provided in the section ``Fused Residual-History Operators'' of the supplementary material.
\fi
The current kernels reflect an initial optimization effort rather than a fully tuned implementation, leaving room for additional speedups through further kernel engineering.

\begin{table}[t]
\centering
\begin{tabular}{lcc}
\toprule
Method & Avg. tok/s (K) & Peak GiB \\
\midrule
Baseline & 114 & 59.8 \\
AttnRes & 104 & 67.4 \\
DAR & 94 & 70.5 \\
\bottomrule
\end{tabular}
\caption{Training efficiency for the 7B sparse-MoE models on one node with eight H800 GPUs. Throughput excludes the first 300 steps to reduce sensitivity to early MoE-routing transients; peak memory is reported per GPU.}
\label{tab:moe-7b-efficiency}
\end{table}

Relative to the residual baseline, Attention Residuals and DAR reduce throughput by 9.16\% and 17.54\%, while increasing per-GPU peak memory by 7.7 and 10.7 GiB, respectively.
Under the matched fused implementations, DAR adds approximately 0.363M total and active parameters over Attention Residuals, reduces throughput by 9.23\%, and uses 3.1 GiB (4.57\%) more peak memory per GPU.
Most of the memory increase already appears with single-stream historical retrieval; DAR's cross-stream retrieval adds a smaller but measurable cost.

\section{Conclusion}

Dual Attention Residuals formulates historical residual retrieval as a joint choice over depth and stream roles.
It uses cross-stream keys to compute historical depth weights and applies them to self-stream values, without changing the Transformer branches.
Across dense and sparse models, DAR improves validation loss.
Together, the ablations and stream analyses show that DAR benefits from dual-stream representation and reciprocal retrieval while using the two streams more complementarily than the tested alternatives.

\ifdefined\DARCombined
\else
\bibliography{custom}
\fi
\DARMaybeEnd

\FloatBarrier
\appendix
\setcounter{secnumdepth}{1}
\counterwithin{figure}{section}
\counterwithin{table}{section}
\counterwithin{equation}{section}
\renewcommand{\theHfigure}{appendix.\thesection.\arabic{figure}}
\renewcommand{\theHtable}{appendix.\thesection.\arabic{table}}
\renewcommand{\theHequation}{appendix.\thesection.\arabic{equation}}
\ifdefined\DARCombined
\let\DARMaybeEnd\relax
\else
\documentclass[letterpaper]{article} 
\usepackage[submission]{aaai2027}  
\usepackage[hyphens]{url}  
\usepackage{graphicx} 
\urlstyle{rm} 
\def\UrlFont{\rm}  
\usepackage{natbib}  
\usepackage{caption} 
\frenchspacing  
\usepackage{booktabs}
\usepackage{amsmath}
\usepackage{amssymb}
\usepackage{array}
\usepackage{listings}
\usepackage[safe]{pbalance}
\usepackage{dblfloatfix}
\definecolor{codeblue}{RGB}{0,0,180}
\definecolor{codegreen}{RGB}{0,115,35}
\definecolor{codered}{RGB}{185,25,25}
\definecolor{codegray}{RGB}{105,105,105}
\lstdefinestyle{darpython}{
  language=Python,
  basicstyle=\ttfamily\scriptsize,
  keywordstyle=\color{codeblue},
  commentstyle=\color{codegreen},
  stringstyle=\color{codered},
  numbers=left,
  numberstyle=\tiny\color{codegray},
  numbersep=7pt,
  frame=single,
  framerule=0.3pt,
  rulecolor=\color{codegray},
  xleftmargin=1.8em,
  framexleftmargin=1.4em,
  columns=fullflexible,
  keepspaces=true,
  showstringspaces=false,
  breaklines=true,
  tabsize=4,
  aboveskip=0pt,
  belowskip=0pt
}
\pdfinfo{
/TemplateVersion (2027.1)
}
\setcounter{secnumdepth}{0}
\setlength{\dblfloatsep}{8pt plus 2pt minus 2pt}

\title{Supplementary Material for Dual Attention Residuals}
\author{Anonymous Submission}
\affiliations{}

\begin{document}
\maketitle
\def\DARMaybeEnd{\end{document}}
\fi

\section{Fused Residual-History Operators}
\label{app:fused-operators}

The 7B sparse-MoE experiments use fused operators, whereas dense runs use direct PyTorch with torch.compile; both implement the same equations and objective.
We retain the main-paper notation and make only token position $p$ explicit; $r$ indexes history states in $\mathcal H_b$, while ``part'' denotes the optional current partial candidate.
The same equations reduce to Full DAR at $K=1$, yielding a layer-wise history.

\paragraph{Block-form pseudocode.}
Figure~\ref{fig:block-dar-pseudocode} gives framework-level pseudocode for Block DAR.
It retains the model logic while omitting fused-kernel and caching optimizations.

\begin{figure*}[b!]
\centering
\begin{minipage}{0.98\textwidth}
\begin{lstlisting}[style=darpython]
def dual_block_attn_res(blocks, partial_block, q0, q1, norm):
    # blocks already include the duplicated token embedding
    states = blocks if partial_block is None else blocks + [partial_block]
    V0 = torch.stack([state[0] for state in states])
    V1 = torch.stack([state[1] for state in states])
    K0, K1 = norm(V1), norm(V0)             # cross-stream keys
    pi0 = softmax(einsum(q0, K0), dim=0)
    pi1 = softmax(einsum(q1, K1), dim=0)
    h0 = einsum(pi0, V0)                    # self-stream values
    h1 = einsum(pi1, V1)
    return alpha_beta_mix(h0, h1)            # branch input h and write gate beta

def mhc_update(partial_block, branch_out, beta, rho_gate):
    write = gated_write(beta, branch_out)
    if partial_block is None:                # first branch in a block
        return write
    rho = rho_gate(partial_block)
    return M(rho) @ partial_block + write     # constrained mHC transport

def forward(self, blocks, hidden_states):
    partial_block = hidden_states
    # apply Block DAR before attention
    h, beta = dual_block_attn_res(
        blocks, partial_block, self.attn_q0, self.attn_q1, self.attn_res_norm)

    # if reaching a block boundary, start a new block
    # block_size counts Transformer layers
    if self.layer_number % self.block_size == 0:
        blocks.append(partial_block)
        partial_block = None

    attn_out = self.attn(self.attn_norm(h))
    partial_block = mhc_update(
        partial_block, attn_out, beta, self.attn_rho_gate)

    # apply Block DAR before MLP
    h, beta = dual_block_attn_res(
        blocks, partial_block, self.mlp_q0, self.mlp_q1, self.mlp_res_norm)
    mlp_out = self.mlp(self.mlp_norm(h))
    partial_block = mhc_update(
        partial_block, mlp_out, beta, self.mlp_rho_gate)
    return blocks, partial_block
\end{lstlisting}
\end{minipage}
\caption{Block DAR pseudocode in the same control flow as Block AttnRes, with cross-stream retrieval and mHC partial-state updates. The variable \texttt{partial\_block} corresponds to $P_b^{i-1}$ in the main paper.}
\label{fig:block-dar-pseudocode}
\end{figure*}

\paragraph{Packed history bank.}
History states $H_0,\ldots,H_{b-1}$ are stored separately from the optional current partial state $P_b^{i-1}$.
At each append, a packing kernel stores the values in $H_r=[v_r^0,v_r^1]$ and their normalized cross-stream key views $\widehat k_r^j=\mathrm{Norm}_h(v_r^{1-j})$; these are only cached views of $\mathcal H_b$.
For token $p$, its depth score is
\begin{equation}
  a_{r\to\ell,p}^{j}
  =(q_\ell^j)^\top \widehat k_{r,p}^j.
\end{equation}

\paragraph{History-bank prefix.}
For every $(\ell,p,j)$, the prefix kernel computes
\begin{align}
  m_{\ell,p}^j
  &=\max_{0\le r<b}a_{r\to\ell,p}^j, \\
  Z_{\ell,p}^j
  &=\sum_{r=0}^{b-1}
    \exp(a_{r\to\ell,p}^j-m_{\ell,p}^j), \\
  U_{\ell,p}^j
  &=\sum_{r=0}^{b-1}
    \exp(a_{r\to\ell,p}^j-m_{\ell,p}^j)v_{r,p}^j.
\end{align}
Here $m_{\ell,p}^j$, $Z_{\ell,p}^j$, and $U_{\ell,p}^j$ are the maximum logit, shifted denominator, and shifted value numerator.
The kernel materializes the logits and max-shifted scores for backward, but not a second tensor of normalized weights; it directly returns
\begin{equation}
  \zeta_{\ell,p}^j=m_{\ell,p}^j+\log Z_{\ell,p}^j,
  \qquad
  \widetilde h_{\ell,p}^{j,\mathrm{hist}}=U_{\ell,p}^j/Z_{\ell,p}^j.
\end{equation}
Here $\zeta_{\ell,p}^j$ and $\widetilde h_{\ell,p}^{j,\mathrm{hist}}$ are the history-bank log-partition and retrieved state.
The matched kernel family pairs the attention and feed-forward queries; DAR additionally evaluates both streams together.

\paragraph{Exact online-softmax merge.}
For branch $i>1$ in block $b$, the additional candidate is $P_b^{i-1}$.
Its cross-stream key, self-stream value, and score are
\begin{align}
  \widehat k_{\mathrm{part},\ell,p}^j
  &=[\mathrm{Norm}_h(P_b^{i-1,1-j})]_p, \\
  v_{\mathrm{part},\ell,p}^j
  &=(P_b^{i-1,j})_p, \\
  a_{\mathrm{part}\to\ell,p}^j
  &=(q_\ell^j)^\top \widehat k_{\mathrm{part},\ell,p}^j.
\end{align}
The scalar
\begin{equation}
  \lambda_{\ell,p}^j
  =\sigma(\zeta_{\ell,p}^j-a_{\mathrm{part}\to\ell,p}^j)
\end{equation}
is the total softmax mass assigned to the history-bank candidates.
Consequently, the residual-read kernel obtains exactly the main paper's retrieval over $\mathcal C_\ell=(\mathcal H_b,P_b^{i-1})$:
\begin{equation}
  h_{\ell,p}^j
  =\lambda_{\ell,p}^j\widetilde h_{\ell,p}^{j,\mathrm{hist}}
   +(1-\lambda_{\ell,p}^j)v_{\mathrm{part},\ell,p}^j.
\label{eq:online-softmax-merge}
\end{equation}
For $i=1$, no partial candidate exists and $h_{\ell,p}^j=\widetilde h_{\ell,p}^{j,\mathrm{hist}}$.
This is the two-phase online-softmax merge of Attention Residuals~\citep{chen2026attentionresiduals}; it is algebraically identical to normalizing all candidates together.

\paragraph{Fusion and backward boundaries.}
The fused residual read performs the exact history--partial merge in Equation~\ref{eq:online-softmax-merge} and predicts $\alpha_\ell$ and $\beta_\ell$.
For $i>1$, it also predicts $\rho_{b,i}$ and returns the transported state $M(\rho_{b,i})P_b^{i-1}$ together with the branch input $h_\ell$.
The unchanged branch $f_\ell$ executes separately; for attention, a post-branch epilogue fuses branch-output handling with the gated write, while the feed-forward path applies the same update directly.
Backward retains the logits, shifted scores, and reduction outputs, but reconstructs packed $v_r^j$ and normalized $\widehat k_r^j$ from the raw $H_r$ states and recomputes current-partial keys as needed.
Its custom kernels differentiate the residual-history path only; the Transformer branches remain outside it and are not recomputed.
After the final block is appended, the learned output queries reuse the same history-bank read without a partial candidate.

\FloatBarrier
\section{Experimental Configurations}
\label{app:experimental-configurations}

Tables~\ref{tab:main-hparams}--\ref{tab:moe-training} report the detailed architecture, optimization, distributed-training, and parameter-count configurations for the dense and sparse-MoE experiments.
Each reported result is obtained from a single training run.

\begin{table*}[t!]
\centering
\small
\setlength{\tabcolsep}{4pt}
\begin{tabular}{@{}>{\raggedright\arraybackslash}p{0.20\textwidth}>{\raggedright\arraybackslash}p{0.25\textwidth}>{\raggedright\arraybackslash}p{0.20\textwidth}>{\raggedright\arraybackslash}p{0.25\textwidth}@{}}
\toprule
Setting & Value & Setting & Value \\
\midrule
Backbone & Dense decoder-only Transformer & Hidden size & 2048 \\
Dataset & FineWeb-Edu-100B & Embedding size & 1024 \\
Training tokens & 50B & Layers & 20 \\
Hardware & $8\times$ H800 GPUs & Attention / KV heads & 16 / 8 \\
GPU nodes & 1 & Feed-forward width & 6656 \\
Training framework & DeepSpeed & Block size $K$ & 2 layers \\
Compilation & torch.compile max-autotune & Optimizer & AdamW \\
Sequence length & 4096 & Peak learning rate & $3.0\times10^{-4}$ \\
Micro / global batch & 2 / 256 & Minimum learning rate & $3.0\times10^{-5}$ \\
Grad. accumulation & 16 & Warmup steps & 2000 \\
ZeRO stage & 1 & Weight decay & 0.1 \\
Activation recompute & Disabled & AdamW $\beta_1,\beta_2$ & 0.9, 0.95 \\
Param./grad. sharding & Disabled & AdamW $\epsilon$ & $1.0\times10^{-8}$ \\
Random seed & 42 & Gradient clipping & 1.0 \\
\bottomrule
\end{tabular}
\caption{Training and architecture details for the 1B language-modeling experiment.
The same optimization recipe is used for all methods in the 1B comparison.}
\label{tab:main-hparams}
\end{table*}

\begin{table*}[t!]
\centering
\small
\setlength{\tabcolsep}{2.5pt}
\begin{tabular}{lcccccccccc}
\toprule
Size & Tokens & Seq. & Batch & Layers & H/KV & $d_{\mathrm{model}}$ & $d_{\mathrm{emb}}$ & $d_{\mathrm{ff}}$ & $K$ & mHC streams \\
\midrule
0.1B & 2B & 4096 & 256 & 16 & 10/5 & 640 & 640 & 2560 & 2 & 4 \\
0.3B & 6B & 4096 & 256 & 20 & 16/4 & 1024 & 1024 & 4096 & 2 & 4 \\
0.5B & 10B & 4096 & 256 & 20 & 20/10 & 1280 & 1024 & 5120 & 2 & 4 \\
\bottomrule
\end{tabular}
\caption{Architecture and training settings for the other dense model sizes.
All runs use FineWeb-Edu-10B, DeepSpeed, torch.compile in \texttt{max-autotune} mode, and random seed 42.
They run on a single node with eight H800 GPUs using micro/global batch sizes 2/256, 16 gradient-accumulation steps, ZeRO Stage~1, no activation recomputation or parameter/gradient sharding, and 300 warmup steps.
Within each size, the residual mechanism changes across methods while the data order, dense decoder backbone, optimization recipe, batch size, and token budget are held fixed.
The column $K$ reports the block size, i.e., the number of Transformer layers per block, for block-form Attention Residuals and block-form DAR; Full Attention Residuals and Full DAR use $K=1$.}
\label{tab:scale-configs}
\end{table*}

\begin{table*}[t!]
\centering
\small
\begin{tabular}{llrr}
\toprule
Nominal scale & Method & Excl. vocabulary (M) & Incl. vocabulary (M) \\
\midrule
0.1B & Baseline & 98.325 & 182.211 \\
0.1B & AttnRes-Block & 98.336 & 182.222 \\
0.1B & AttnRes-Full & 98.346 & 182.232 \\
0.1B & DAR-Block & 98.449 & 182.335 \\
0.1B & DAR-Full & 98.571 & 182.457 \\
0.1B & mHC-4 & 100.292 & 184.178 \\
\midrule
0.3B & Baseline & 304.129 & 438.347 \\
0.3B & AttnRes-Block & 304.151 & 438.368 \\
0.3B & AttnRes-Full & 304.170 & 438.388 \\
0.3B & DAR-Block & 304.377 & 438.595 \\
0.3B & DAR-Full & 304.621 & 438.839 \\
0.3B & mHC-4 & 308.062 & 442.280 \\
\midrule
0.5B & Baseline & 491.572 & 659.345 \\
0.5B & AttnRes-Block & 491.599 & 659.372 \\
0.5B & AttnRes-Full & 491.624 & 659.396 \\
0.5B & DAR-Block & 491.882 & 659.655 \\
0.5B & DAR-Full & 492.187 & 659.959 \\
0.5B & mHC-4 & 496.489 & 664.261 \\
\midrule
1B & Baseline & 1{,}069.631 & 1{,}338.067 \\
1B & AttnRes-Block & 1{,}069.674 & 1{,}338.110 \\
1B & DAR-Block & 1{,}070.127 & 1{,}338.563 \\
\bottomrule
\end{tabular}
\caption{Exact dense-model parameter counts. ``Excl.'' and ``incl.'' indicate whether vocabulary parameters are excluded or included. The scale labels in the main paper refer to the nominal non-vocabulary model size.}
\label{tab:dense-parameter-counts}
\end{table*}

The sparse backbone uses grouped-query attention~\citep{ainslie2023gqa}, RoPE~\citep{su2024roformer}, RMSNorm~\citep{zhang2019rmsnorm}, SwiGLU~\citep{shazeer2020glu}, and sparsely gated mixture-of-experts routing~\citep{shazeer2017moe}.

\begin{table*}[t!]
\centering
\small
\setlength{\tabcolsep}{3pt}
\begin{tabular}{@{}p{0.21\textwidth}p{0.23\textwidth}p{0.21\textwidth}p{0.23\textwidth}@{}}
\toprule
Architecture setting & Value & Architecture setting & Value \\
\midrule
Transformer layers & 24 & Context length & 4096 \\
Hidden size & 2048 & Expert FFN width & 704 \\
Attention heads / KV groups & 16 / 8 & Vocabulary size & 131{,}072 \\
Position encoding & RoPE & RoPE base & $10^6$ \\
Normalization / activation & RMSNorm / SwiGLU & Linear biases & Disabled \\
Attention backend & Fused & Input--output embeddings & Tied \\
Block size $K$ & 3 layers & Experts & 64 \\
Experts selected per token & 8 & Router balancing & Auxiliary loss \\
Auxiliary-loss coefficient & 0.01 & Expert computation & Grouped GEMM \\
Token dispatcher & All-to-all & & \\
\bottomrule
\end{tabular}
\caption{Architecture of the sparse-MoE backbone. The residual baseline, Attention Residuals, and DAR share this backbone, the pretokenized FineWeb-Edu-100B data, optimization schedule, and distributed-training configuration. Attention Residuals and DAR use block size $K=3$; the residual mechanism and its matched fused implementation are the only method-specific components. The expert FFN width is the intermediate width of each routed expert.}
\label{tab:moe-architecture}
\end{table*}

\begin{table*}[t!]
\centering
\small
\begin{tabular}{lrrrr}
\toprule
Method & Total excl. vocab. (B) & Total incl. vocab. (B) & Active excl. vocab. (B) & Active incl. vocab. (B) \\
\midrule
Baseline & 6.949 & 7.217 & 1.136 & 1.404 \\
AttnRes & 6.949 & 7.217 & 1.136 & 1.404 \\
DAR & 6.949 & 7.218 & 1.136 & 1.405 \\
\bottomrule
\end{tabular}
\caption{Exact sparse-MoE parameter counts. Active parameters are those used per token under top-8 expert routing; vocabulary parameters are reported separately because they contribute equally to all methods.}
\label{tab:moe-parameter-counts}
\end{table*}

\begin{table*}[t!]
\centering
\small
\setlength{\tabcolsep}{3pt}
\begin{tabular}{@{}p{0.21\textwidth}p{0.23\textwidth}p{0.21\textwidth}p{0.23\textwidth}@{}}
\toprule
Training setting & Value & Training setting & Value \\
\midrule
Dataset & FineWeb-Edu-100B (pretokenized) & Precision & BF16 \\
Training framework & Megatron-Core & Activation recomputation & Selective \\
Training iterations & 60{,}000 & Training tokens & 31.46B \\
Micro / global batch size & 2 / 128 & Hardware & 1 node / 8 H800 GPUs \\
Optimizer & AdamW & Weight decay & 0.1 \\
AdamW $\beta_1,\beta_2$ & 0.9, 0.95 & AdamW $\epsilon$ & $10^{-8}$ \\
Peak / minimum learning rate & $3\times10^{-4}$ / $3\times10^{-5}$ & Gradient clipping & 1.0 \\
Warmup & 2{,}000 iterations & Decay schedule & Cosine over 60{,}000 iterations \\
Distributed optimizer & Enabled & Sequence parallelism & Enabled \\
TP / PP / EP / CP & 2 / 1 / 4 / 1 & Grad. accumulation & 16 \\
Data-loader mode & Cyclic & Random seed & 42 \\
\bottomrule
\end{tabular}
\caption{Training and distributed-system configuration for the sparse-MoE comparison. TP, PP, EP, and CP denote tensor, pipeline, expert, and context parallelism. On eight GPUs, TP${}=2$ gives data-parallel size 4; micro-batch size 2 with 16 accumulation steps yields global batch 128. At context length 4096, 60{,}000 iterations correspond to 31.46B training tokens.}
\label{tab:moe-training}
\end{table*}

\FloatBarrier
\section{Detailed 1B Evaluation}
\label{app:1b-evaluation}

The three 1B checkpoints are evaluated once under identical protocols.
Paloma~\citep{magnusson2024paloma} uses its official evaluation implementation; Table~\ref{tab:paloma-domains} reports token perplexity for each evaluated domain group.
For source $s$ with $N_s$ scored tokens, the aggregate token perplexity is
\begin{equation}
\mathrm{PPL}_{\mathrm{token}}
=
\exp\!\left(
-\frac{\sum_{s=1}^{16}\sum_{t=1}^{N_s}\log p(x_{s,t})}
{\sum_{s=1}^{16}N_s}
\right).
\label{eq:paloma-token-ppl}
\end{equation}
Byte PPL and BPB use the same pooled log-likelihood normalized by the total number of bytes, with exponential and base-2 conversions, respectively.
Thus, all aggregate metrics are token- or byte-level micro averages over the 16 sources, rather than macro averages of their source-level perplexities.
The zero-shot results in the main paper use the default task formulations in lm-evaluation-harness~\citep{biderman2024reproducible}.
The reported metric is length-normalized accuracy (\texttt{acc\_norm}) for HellaSwag, PIQA, ARC-Easy, OpenBookQA, SocialIQA, and SciQ, and accuracy (\texttt{acc}) for WinoGrande and BoolQ.
Because each checkpoint is evaluated once, the paper reports point estimates without variance or significance claims.

\begin{table}[h!]
\centering
\small
\setlength{\tabcolsep}{2.2pt}
\begin{tabular}{@{}lrrr@{}}
\toprule
Domain group & Baseline & AttnRes & DAR \\
\midrule
4chan & 631.49 & \textbf{518.60} & 536.45 \\
C4 100 Domains & 37.56 & 37.16 & \textbf{35.98} \\
C4 & 44.04 & 43.75 & \textbf{42.33} \\
Dolma v1.5 & 102.37 & 96.63 & \textbf{93.23} \\
100 Programming Langs. & 41{,}241.97 & 24{,}985.74 & \textbf{23{,}731.25} \\
100 Subreddits & 87.58 & 86.48 & \textbf{84.12} \\
Falcon RefinedWeb & 51.08 & 50.27 & \textbf{48.79} \\
Gab & 8{,}203.23 & 7{,}966.30 & \textbf{7{,}435.71} \\
M2D2 S2ORC & 52.05 & 49.84 & \textbf{48.26} \\
M2D2 Wikipedia & 47.63 & 46.86 & \textbf{45.35} \\
Manosphere & 94.71 & 93.53 & \textbf{91.01} \\
mC4 & 76.08 & 73.76 & \textbf{71.39} \\
PTB & 53.38 & 50.92 & \textbf{47.78} \\
RedPajama & 590.75 & 508.56 & \textbf{501.21} \\
Twitter AAE & 8{,}396.23 & 8{,}083.35 & \textbf{7{,}606.84} \\
Wikitext-103 & 28.06 & 26.86 & \textbf{26.31} \\
\bottomrule
\end{tabular}
\caption{Token perplexity across the 16 reported Paloma domain groups for the 1B models. Lower is better. DAR is best on 15 groups; Attention Residuals is best on 4chan.}
\label{tab:paloma-domains}
\end{table}

\FloatBarrier
\section{Additional Analyses}
\label{app:additional-analyses}

We report the router-balance and historical retrieval diagnostics referenced in the main paper.

\begin{figure}[h!]
\centering
\includegraphics[width=\columnwidth]{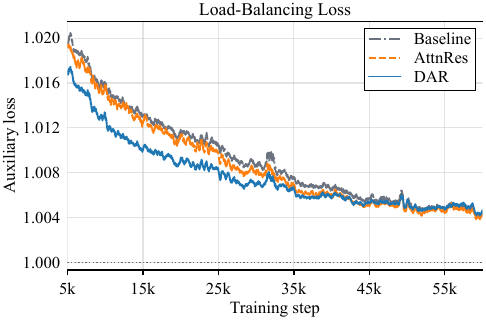}
\caption{All three sparse-MoE models converge toward balanced aggregate expert utilization. The unscaled router auxiliary loss specified in Table~\ref{tab:moe-architecture} is shown from 5K to 60K steps using a 300-step moving average. Under the adopted normalization, the dotted reference at 1.0 denotes uniform aggregate utilization; the closely matched curves show that the residual mechanism does not materially alter router balance.}
\label{fig:moe-load-balancing}
\end{figure}

\begin{figure}[t!]
\centering
\includegraphics[width=\columnwidth]{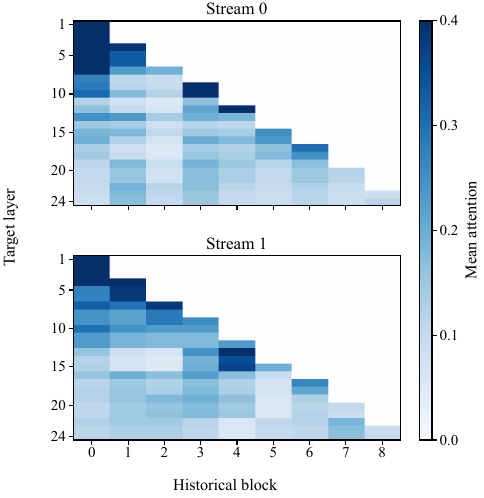}
\par\smallskip
\centerline{\small (a) Pre-attention retrieval}
\medskip
\includegraphics[width=\columnwidth]{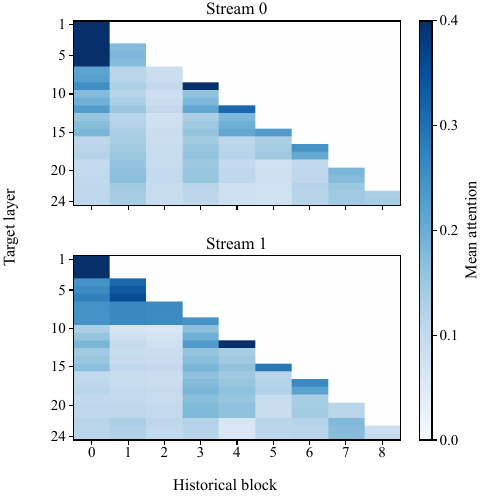}
\par\smallskip
\centerline{\small (b) Pre-MLP retrieval}
\caption{Historical retrieval weights for 7B sparse-MoE DAR, averaged over eight 4096-token sequences. Rows are target layers and columns are historical blocks; blank cells are unavailable. Both branch types exhibit the same stream separation.}
\label{fig:moe-7b-retrieval-maps}
\end{figure}

\begin{figure*}[t!]
\centering
\includegraphics[width=0.88\textwidth]{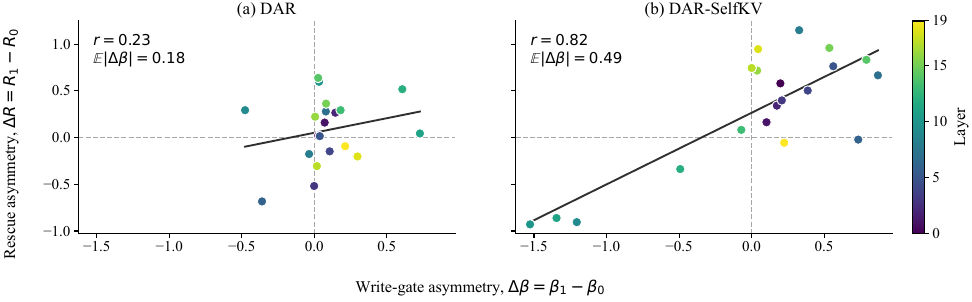}
\caption{DAR-SelfKV exhibits stronger write-gate asymmetry and a stronger gate--rescue association than DAR. Each point is one MLP layer, colored by depth; lines are least-squares fits and both panels use identical axes.}
\label{fig:selfkv-gate-rescue}
\end{figure*}

\paragraph{Gate asymmetry protocol.}
For Figure~\ref{fig:selfkv-gate-rescue}, we instrument the trained 0.5B DAR and DAR-SelfKV models on four FineWeb-Edu-10B validation sequences of length 4096 and average their MLP write gates over the final 128 token positions.
At Transformer layer $n$, let $\beta_n^j$ be the post-MLP branch-write gate for stream $j$ and $R_{n,j}$ its rescue gain.
We define the write and rescue gaps as
\[
  g_n=\mathbb{E}_{x,t}[\beta_n^1-\beta_n^0],
  \qquad
  d_n=R_{n,1}-R_{n,0},
\]
where the expectation averages over validation sequences and token positions.
We omit the residual keep gate $\rho$ because it enters the symmetric two-stream mixing matrix and does not by itself favor either stream.
DAR-SelfKV has a larger mean absolute write gap than DAR ($0.49$ versus $0.18$), and its write and rescue gaps are more strongly correlated across layers ($r=0.82$ versus $0.23$).
Thus, SelfKV's high cross-stream similarity masks imbalanced information flow: one stream receives stronger writes and becomes functionally dominant, whereas DAR uses both trajectories more evenly.

\ifdefined\DARCombined
\else
\FloatBarrier
\begingroup
\raggedbottom
\bibliography{custom}
\endgroup
\fi
\DARMaybeEnd

\FloatBarrier
\begingroup
\raggedbottom
\bibliography{custom}
\endgroup

\end{document}